\pdfoutput=1

\documentclass[11pt]{article}

\usepackage{acl}

\usepackage{times}
\usepackage{latexsym}
\usepackage{float}
\usepackage{graphicx}
\usepackage{subcaption}

\usepackage[T1]{fontenc}

\usepackage[utf8]{inputenc}

\usepackage{microtype}

\usepackage{inconsolata}
\usepackage{titlesec}

\titleformat{\paragraph}
{\normalfont\normalsize\bfseries}{\theparagraph}{1em}{}
\titlespacing*{\paragraph}
{0pt}{3.25ex plus 1ex minus .2ex}{1.5ex plus .2ex}

\title{Verb Categorisation for Hindi Word Problem Solving}

\author{Harshita Sharma, Pruthwik Mishra, Dipti Misra Sharma \\
        IIIT-Hyderabad\\ \{harshita.sharma, pruthwik.mishra\}@research.iiit.ac.in, dipti@iiit.ac.in}

\begin{document}

\maketitle
\begin{abstract}
Word problem Solving is a challenging NLP task that deals with solving mathematical problems described in natural language. Recently, there has been renewed interest in developing word problem solvers for Indian languages. As part of this paper, we have built a Hindi arithmetic word problem solver which makes use of verbs. Additionally, we have created verb categorization data for Hindi. Verbs are very important for solving word problems with addition/subtraction operations as they help us identify the set of operations required to solve the word problems. We propose a rule-based solver that uses verb categorisation to identify operations in a word problem and generate answers for it. To perform verb categorisation, we explore several approaches and present a comparative study. 
\end{abstract}

\section{Introduction}


Verb Categorisation is the most intuitive and explainable semantic parsing approach for word problem solving. This approach was introduced in \citet{hosseini-etal-2014-learning}. It uses verbs to identify operations 
required to solve a word problem. The idea is to identify the following parts of a word problem on top of which we use verb categories to perform calculations:
\begin{itemize}
    \item \textbf{Entities}: Objects whose quantity is observed or updated through the course of the word problem.
    \item \textbf{Attributes of Entities}: A characteristic quality or feature of an entity. These are usually marked by adjectives.
    \item \textbf{Containers}: A container refers to a group of entities. It may refer to any animate/inanimate object that possesses or contains entities.
    \item \textbf{Quantities of Entities}: The number of entities in a given container. Quantities in a container can be known or unknown.
\end{itemize}

This can be explained using the following example:
\begin{figure}[h]
  \centering
  \includegraphics[width=1\linewidth]{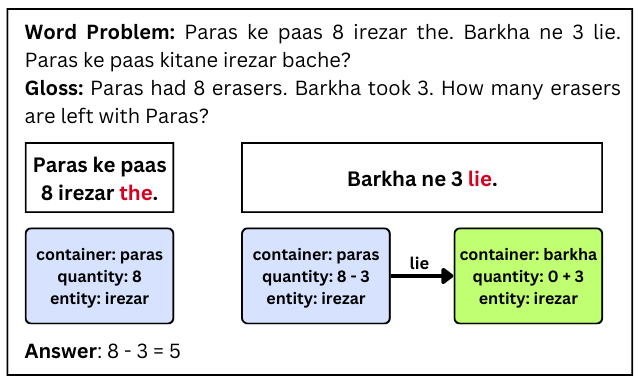}
  \caption{Example of solving word problem using verbs}
  \label{fig:basic-eg}
\end{figure}

Leveraging the insights provided by \citet{hosseini-etal-2014-learning}, as part of this paper, we make the following contributions:
\begin{itemize}
    \item Redefine verb categories for word problem solving.
    \item Create verb categorisation data for the Hindi language.
    \item Introduce three new verb categorisation approaches and provide a comprehensive comparative analysis of these approaches.
    \item Introduce a rule-based solver\footnote{Code and data can be found here: \url{https://github.com/hellomasaya/verb-cat-for-hindi-wps}} that uses verbs to identify specific mathematical operations to solve word problems.
\end{itemize}

\section{Motivation}
\begin{figure}[h]
  \centering
  \includegraphics[width=1\linewidth]{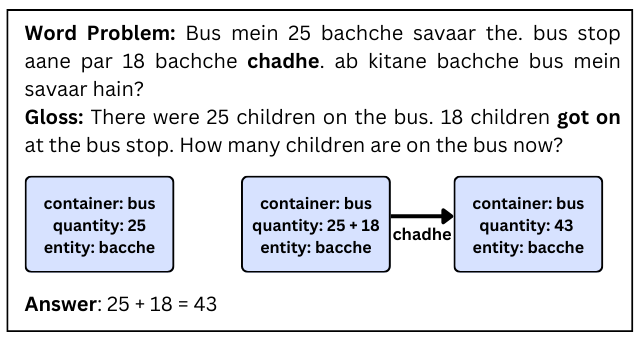}
  \caption{Solving word problem with verb - 1}
  \label{fig:with-verb}
\end{figure}

\begin{figure}[h]
  \centering
  \includegraphics[width=1\linewidth]{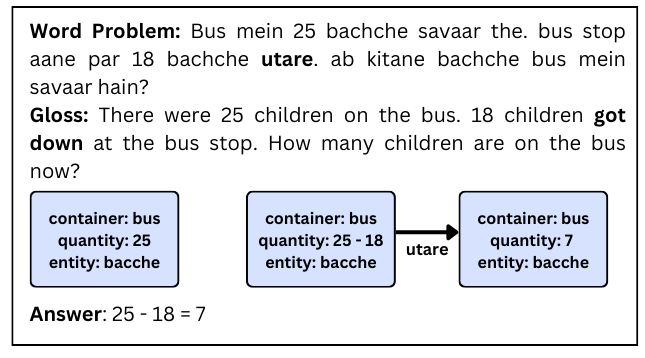}
  \caption{Solving word problem with verb - 2}
  \label{fig:with-verb-2}
\end{figure}

Let us take an example to understand the vital role of verbs in solving a word problem as shown in the following figures. Figure~\ref{fig:no-verb} shows a word problem with containers, entities, and their quantities by masking the verbs. Here, we cannot identify the operation needed to reach the final state and answer the question asked in the word problem. However, in Figure \ref{fig:with-verb}, we are able to reach the final state. Moreover, when we change the verb from Figure~\ref{fig:with-verb} to Figure~\ref{fig:with-verb-2}, the operation also changes.

\begin{figure}[ht]
  \centering
  \includegraphics[width=1\linewidth]{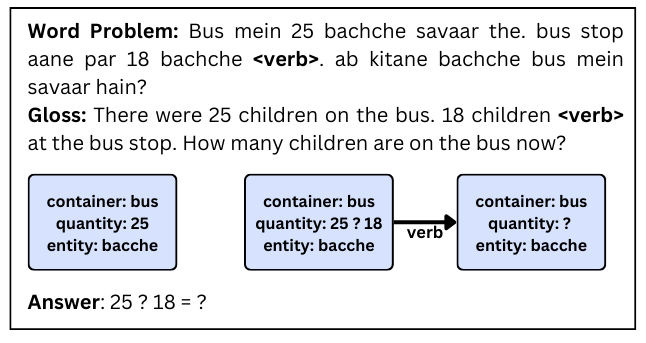}
  \caption{Example of solving word problem without verb}
  \label{fig:no-verb}
\end{figure}

\begin{table*}
	\centering
	\begin{tabular}{|p{3cm}|p{12cm}|} 
		\hline
		\textbf{Verb Category} & \textbf{Definition} \\
		\hline
		Observation & It states just the presence of entities in a container  \\
   		\hline
		Positive & It states the number of entities being added to a container or which are created in a container.\\
\hline
		Negative & It states the quantity of entities being removed or destroyed from a container.\\
		 \hline
   		Positive Transfer & It is associated with statements that involve two containers. It states a transfer of the number of entities from the second container to the first.\\
		 \hline
   		Negative Transfer & It is associated with statements that involve two containers. It states a transfer of the quantity of entities from the first container to the second.\\
		\hline
	\end{tabular}
	\caption{Five Verb Categories}
	\label{table: Five Verb Categories}
\end{table*}
\section{Verb Categorisation}
This section focuses on the first step of word problem solving using verb categorisation, i.e. classifying verbs into semantic categories. Since verbs can be used to identify only positive and negative operations, we filtered the HAWP dataset\footnote{\url{https://github.com/hellomasaya/hawp}} \cite{sharma-etal-2022-hawp} to have only word problems involving addition and subtraction operations. Verbs tell us whether entities are observed, created, destroyed, or transferred. For multiplication and division, we need another layer of categorisation with different Part of Speech categories on top of verb categorisation.

\subsection{Verb Categories}
Table~\ref{table: Five Verb Categories} lists the five categories we have used in this paper. We also included a sixth category - \textit{na}. During POS tagging, non-verbs were tagged as verbs; these tokens were put into the \textit{na} category.

\subsection{Annotation of Verbs}
In the HAWP dataset (2336 word problems), 1713 word problems are based on addition and subtraction operations. These problems feature in our dataset for word problem solving using verb categorisation.
In these 1713 word problems, there are around 200 unique verbs. These verbs were annotated with the categories mentioned above. 

\citet{hosseini-etal-2014-learning} have seven verb categories that are container-centric. They have two additional categories of \textit{Construct} and \textit{Destroy} apart from the ones defined in Table~\ref{table: Five Verb Categories}. But these two resemble \textit{Positive} and \textit{Negative} categories, respectively. Hence, we decide to drop these two categories. 

For the verb annotation task, two annotators with post-graduate levels of education in computational linguistics are involved. We conduct experiments to evaluate the inter-annotator agreement between them on 225 verbs from 100 sentences. The Fleiss'\footnote{\url{https://en.wikipedia.org/wiki/Fleiss\%27\_kappa}}
kappa score of agreement is 0.89, which denotes almost perfect agreement. There was maximum disagreement between \textit{Observation} and \textit{Positive} classes.


\subsection{Approaches}

We tried mainly three kinds of approaches, which are detailed in the following subsections. All the approaches are evaluated using a 5-fold cross-validation technique.
\subsubsection{Verb Distance}
The first approach is the training less method using verb distance. Each verb in Hindi is represented by its pre-trained FastText word vector \cite{grave-etal-2018-learning} of 300 dimensions. A test verb is assigned the verb category corresponding to its closest training verb. We implemented this approach using the 1-nearest neighbour approach.


\subsubsection{Statistical Models}
\begin{figure}[h]
  \centering
  \includegraphics[width=1\linewidth]{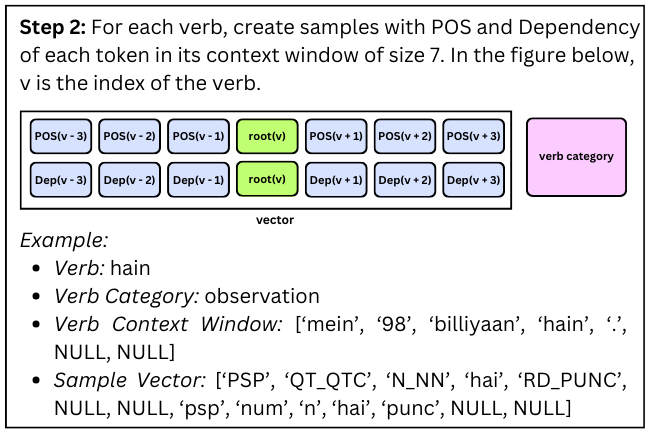}
  \caption{Step 2 of Data Preparation}
  \label{fig:stat-2}
\end{figure}

\paragraph{Data Preparation}
The idea is to use a bag of words representation for a verb and its neighbours in their actual order as a sample and the category of the verb as the label.  We created samples for the task using word-level information as indicated in Figure~\ref{fig:stat-2}. After trying context windows of different sizes, we finalised the context window size as 7. 
\begin{figure}[h!]
  \centering
  \includegraphics[width=1\linewidth]{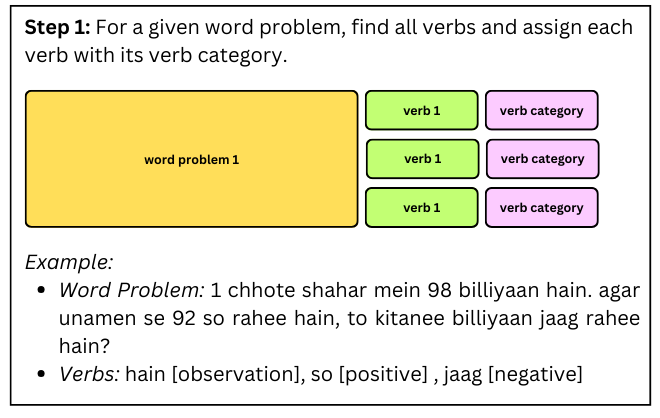}
  \caption{Step 1 of Data Preparation}
  \label{fig:stat-1}
\end{figure}
Therefore, we will have word-level information for three neighbours to the right of the verb and the same for three neighbours to the left of the verb. We parse each sentence using an in-house shallow parser \cite{mishra2023multi} for identifying the POS tag and root of each word. We used ISC-parser from Natural language tool-kit for Indian Language Processing \footnote{\url{https://github.com/iscnlp/iscnlp/tree/master/iscnlp}} to get dependency tags of each token in each sentence of each word problem in the dataset.

\paragraph{Experimental Setup}
We performed this classification task using 3 machine-learning approaches:
        \begin{itemize}
            \item Logistic Regression
            \item Random Forest
            \item Support Vector Machines (SVM)
        \end{itemize}

All these models have been implemented using Scikit-learn \cite{scikit-learn} machine learning framework.
\begin{figure}[h!]
  \centering
  \includegraphics[width=1\linewidth]{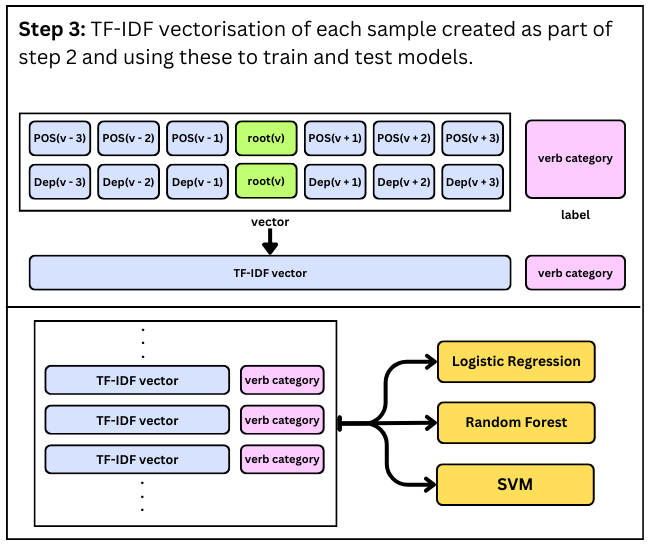}
  \caption{Overview of Training Statistical Models}
  \label{fig:stat-3}
\end{figure}


\subsubsection{MuRIL Contextual Embeddings}
Contextual embeddings, especially BERT \cite{devlin-etal-2019-bert} based embeddings, have been shown to be very effective for classification and generalization tasks. BERT is trained in two stages: pre-training and fine-tuning. The model is first trained on a huge monolingual corpus to learn language-specific representations and then fine-tuned on a downstream task. In our case, the downstream task is the verb categorization task. As this is a text or sentence classification task, it is a perfect test for using BERT or BERT-like models. For this, we used MuRIL \cite{khanuja2021muril}, a multilingual transformer \cite{vaswani2017attention} model trained on English and 16 Indian languages. MuRIL is pre-trained using masked language modelling as well as translation language modelling. It has a combined vocabulary of 197K words.
\paragraph{Data Preparation}
We used the 1713 word problems from HAWP. Since MuRIL can handle large contexts, we do not limit ourselves to a fixed context window. For this task, all the words till a verb is encountered constitute a sample. A total of 6506 samples were created for verb categorization. Let us take an example to understand this better.
\begin{itemize}
    \item Original Question:\\
kanishk ko samudr tat par 47 seepiyaan mileen, usane laila ko 25 seepiyaan deen. usake paas ab kitanee seepiyaan hain?

    \item [] Gloss: Kanishk found 47 shells on the beach, he gave 25 shells to Laila. How many shells does he have now?
    \item Samples for Verb Categorization
    \begin{itemize}
        \item kanishk ko samudr tat par 47 seepiyaan \underline{mileen}
        \item [] Gloss: Kanishk (found) 47 shells on the beach
        \item kanishk ko samudr tat par 47 seepiyaan mileen, usane laila ko 25 seepiyaan \underline{deen}.
        \item [] Gloss: Kanishk found 47 shells on the beach, he (gave) 25 shells to Laila.
        \item 
            usake paas ab kitanee seepiyaan \underline{hain}?
        \item [] Gloss: How many shells does he (have) now?
    \end{itemize}
\end{itemize}
\paragraph{Experimental Setup}
MuRIL has 236 million parameters, and it uses AdamW \cite{loshchilov2017decoupled} optimizer. We use a 5-fold cross-validation technique to evaluate the model. MuRIL is fine-tuned for ten epochs with a batch size of 4. MuRIL-based text classification model is implemented using HuggingFace \cite{wolf2019huggingface} library.

\subsection{Results and Discussion}
The results from all models are shown in Table~\ref{tab:muril}.

\begin{table}[h]
    \centering
    \begin{tabular}{l|l}\hline
       \textbf{Approach} & \textbf{F1-score} \\ \hline
        Verb Distance & 0.895\\\hline
        Logistic Regression & 0.865 \\\hline
 Random Forest & 0.883 \\\hline
 Support Vector Machines & 0.904 \\\hline
        MuRIL Fine-tuning & \textbf{0.962}\\\hline
    \end{tabular}
    \caption{Verb Categorization Results with different approaches}
    \label{tab:muril}
\end{table}
We can observe that MuRIL Fine-tuning outperforms other approaches by a significant margin. The class \textit{na} contains the highest classification error. The major cause of ambiguity is between the \textit{Observation} and \textit{Positive} in all the models.
\section{Solver}
We build a simple rule-based system that takes in a word problem and generates answers to the word problem.
For each problem, we iterate through all tokens in each of its sentences and follow the rules mentioned below.
\subsection{Find container, quantity and entity}\label{subsec:cqe}
\begin{itemize}
    \item A container is a \textit{proper noun} or \textit{adverb of place}. 
    \item A quantity is a \textit{number}. Whenever a quantity is found, the last identified container is associated with this quantity. 
    \item An entity is a \textit{noun}. If there is an adjective associated with this entity, it is clubbed with the entity. When a word problem has the \textit{Rupee} symbol, the entity is taken to be this symbol itself. Examples of these rules can be found in Figures~\ref{fig:transfer-verb-eg} and \ref{fig:neg-verb-eg}.
\end{itemize}

\subsection{Store States}\label{solver:store-states}
Here, a state refers to the status of an entity that stores information about an entity, its container, its associated quantity, and any attributes of the entity.
\begin{enumerate}
    \item Once an entity is found, the associated quantity and container are used to form a state.
    \item Before storing the quantity in a state, if the verb that follows the identified entity of this state has its verb category as Negative, the quantity is negated and stored. A detailed example of applying these rules can be found in Figure~\ref{fig:neg-verb-eg}.
\end{enumerate}

\subsection{Handle Transfer Category}\label{subsec:handle-transfer}
\begin{figure*}[h]
\centering
  \includegraphics[width=0.97\linewidth]{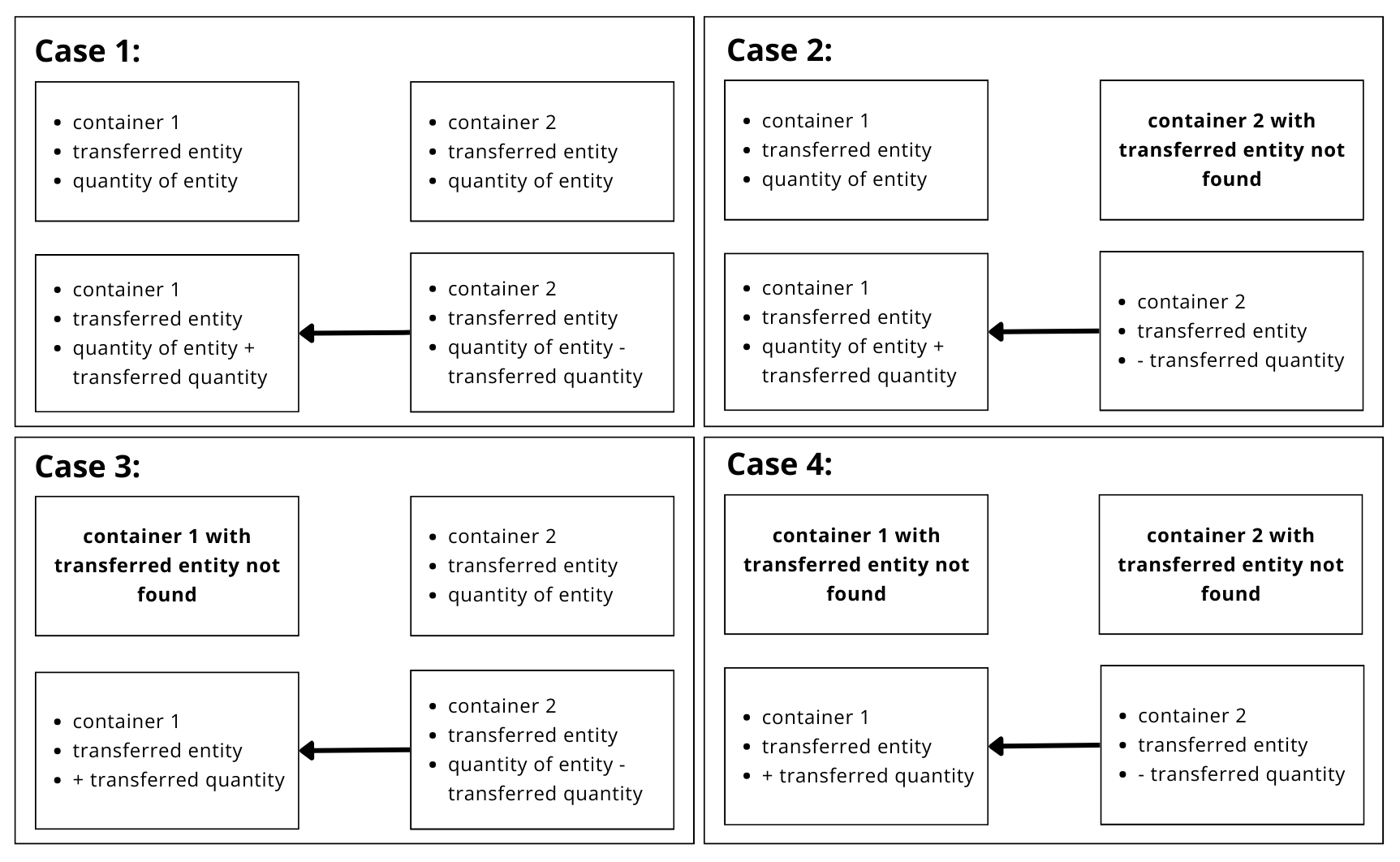}
  \caption{Transfer for verb category: Positive Transfer}
  \label{fig:sub1}
\end{figure*}

\begin{figure*}[ht]
  \centering
  \includegraphics[width=0.97\linewidth]{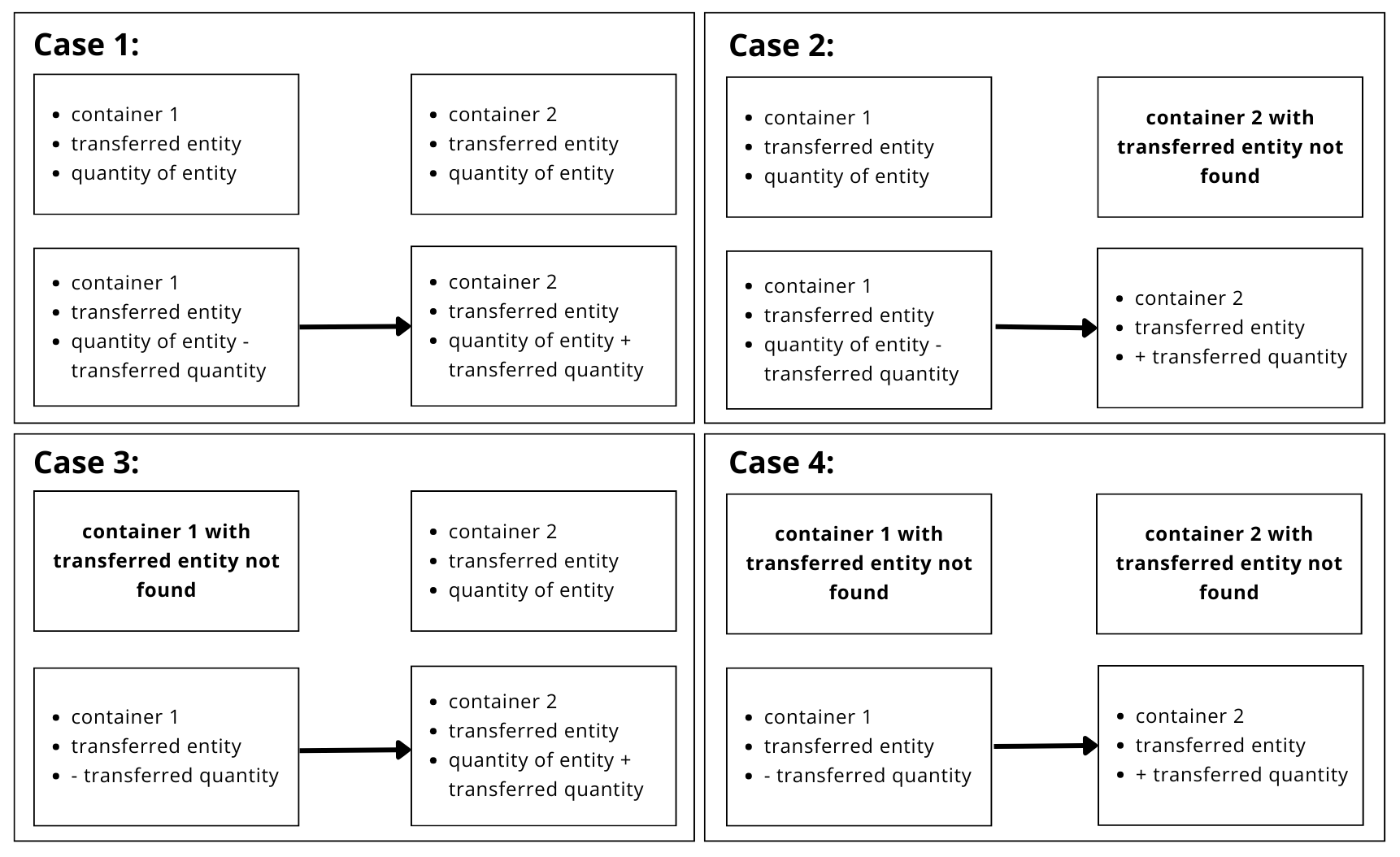}
  \caption{Transfer for verb category: Negative Transfer}
  \label{fig:sub2}
\end{figure*}

\begin{enumerate}
    \item Once a verb is found in the word problem, we check for Transfer categories. We check if the verb belongs to the Positive Transfer or the Negative Transfer category from our verb categorisation exercise.
    \item If the Transfer verb category is found, we find transfer components, i.e. transfer containers (two containers b/w which transfer is taking place) and the quantity of entities being transferred.
    \item Then, we iterate through the states and find which already present states have transfer-container 1 and transfer-container 2. Then, we check if transfer-entity is present in these states.
    \item Finally, we update the previous states based on which containers and entities are already available in the previous states. These cases for positive and negative categories can be seen in Fig \ref{fig:sub1} and Fig \ref{fig:sub2}. A detailed example of applying these rules can be found in Fig \ref{fig:transfer-verb-eg}.
\end{enumerate}


\subsection{Finding Answer}\label{subsec:answer}
\begin{figure*}[h]
  \centering
  {%
\setlength{\fboxsep}{0pt}%
\setlength{\fboxrule}{1pt}%
\fbox{\includegraphics[width=1\linewidth]{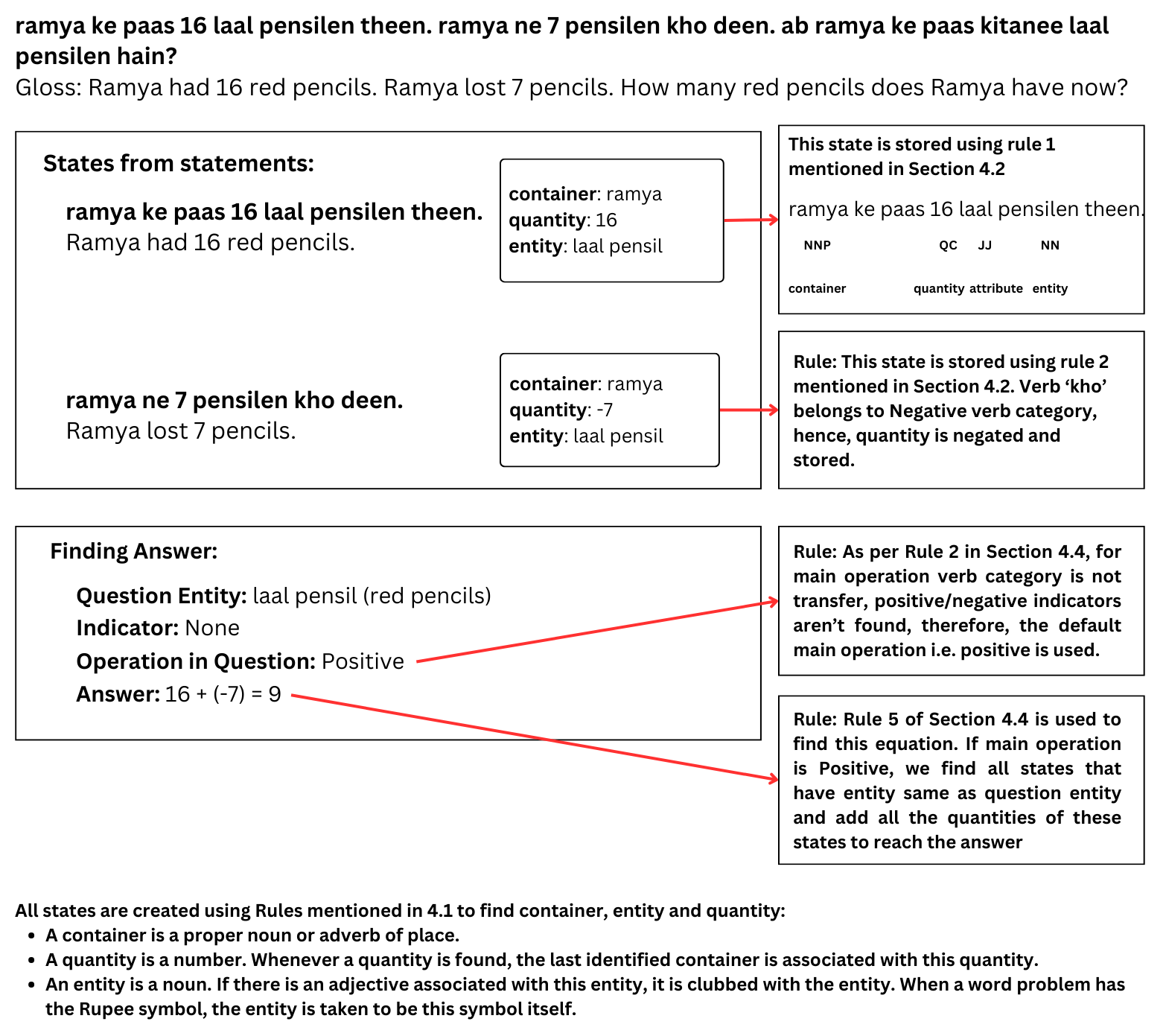}}}
  \caption{Example of solving word problem using negative verb category}
  \label{fig:neg-verb-eg}
\end{figure*}

\begin{enumerate}
    \item Find Question Entity (and Question Container, only in the case when a transfer verb is encountered) from the question using the same rules mentioned in Section \ref{subsec:cqe}.
    \item Find Main Operation.
    \begin{itemize}
        \item If a transfer verb category is encountered in the word problem, the main operation is Transfer.
        \item If any positive indicator is present in question, the main operation is Positive.
        \item If any negative indicator is present in question, the main operation is negative.
        \item The main operation is positive if none of the above conditions are met.
    \end{itemize}
    \begin{table}[h]
    \centering
    \begin{tabular}{p{1.75cm}|p{4cm}}\hline
    \textbf{Indicators} & \textbf{Examples} \\\hline
        Positive Indicators & ‘kul’, ‘milakar’, ’milkar’ etc.\\\hline
        Negative Indicators & ‘mukable’, ‘tulna’, ‘pehle’, ‘chahiye’ etc. \\\hline
    \end{tabular}
    \caption{Indicators in questions}
    \label{tab:indicators}
\end{table}
\item If the main operation is Transfer, our calculation is already complete as part of \ref{subsec:handle-transfer}. We find the state that has the answer to the question by looking at all the states we created, and whichever state matches the question’s container and entity pair, we return its quantity as the answer. A detailed example of the transfer verb category is explained in Fig \ref{fig:transfer-verb-eg}. More examples can be found in the Appendix.

\item If the main operation is Negative, we find all states that have the same entity as the question entity. Then, keeping the quantity in the first state as it is, we subtract the quantities of the states that follow from it to finally reach the answer.
\item If the main operation is Positive, we find all states that have the same entity as the question entity and add all the quantities of these states to finally reach the answer. A detailed example of this case is explained in Fig \ref{fig:neg-verb-eg}.
\end{enumerate}

\subsection{More Rules}\label{subsec:more-rules}
\begin{enumerate}
    \item If the final answer calculated by the solver is negative, we return its absolute value. 
    \item While identifying relevant entities from states, if the entities from the question and a state match but the attribute is missing in either state or question, we still regard the entity as relevant.
    \item If the entity in the word problem is found to be one of ‘paisa’, ‘keemat’, ‘laagat’, and ‘rupay’, we change it to the \textit{Rupee} symbol. 
    \item If a quantity is found without an entity or container, we retain the same entity and container from the last state and create a new state with the quantity found. This is called circumscription assumption \cite{mccarthy1980circumscription}.
    \item If the entity in question is not found in states, we assume the entity of the first state to be the entity in the question and perform the steps for finding the answer.
\end{enumerate}

Detailed examples of all rules can be found in the Appendix.

\subsection{Results and Discussion}
The solver was tested on test sets using predicted verb categories, and an average accuracy of 41.2\% was reported, which is comparable to 40.04\% reported by \citet{sharma-etal-2022-hawp} for one-operation problems in the HAWP dataset. Some of the cases in which the solver fails are listed below:
\begin{itemize}
    \item Irrelevant Information: The solver fails to identify some cases of irrelevant information. 
    \item Error in entity/container/action identification.
    \item Set Completion: The solver fails to handle word problems which require the knowledge of set completion. 
    \item Parsing Errors: Errors caused by incorrectly tagged part of speech. This also includes cases when parsers miss foreign words. 
    \item Rules: There are cases when a rule that works for some examples may not work for others. 
\end{itemize}

Examples of these cases can be found in the Appendix in Table \ref{table:solver-errors}.

\section{Limitations}
Apart from the limitations of the solver, the method of using verb categorisation to solve word problems also has some limitations. As stated, solving word problems using verb categorisation is only limited to addition and subtraction word problems because verbs can only help us identify these operations. Moreover, there were errors in the dependency labels. Since verb categorisation very heavily relies on these parsers for finding verb categories and identifying entities, containers and actions/verbs for solving word problems. This adds to the limitation of this method.

\section{Conclusion and Future Work}
In this paper, we create a rule-based and easily explainable solver that uses a verb categorisation technique to identify operations to solve word problems. This can be used as a teaching aid for both students and teachers. As part of the verb categorisation task, we run experiments with three approaches: Verb Distance (no training involved), statistical, and neural approaches using MuRIL. As part of future work, we will explore more approaches to improve the accuracy of our solver and its range, i.e., solving word problems with multiplication and division operations.


\begin{figure*}[h]
  \centering
  {%
\setlength{\fboxsep}{0pt}%
\setlength{\fboxrule}{1pt}%
\fbox{\includegraphics[width=1\linewidth]{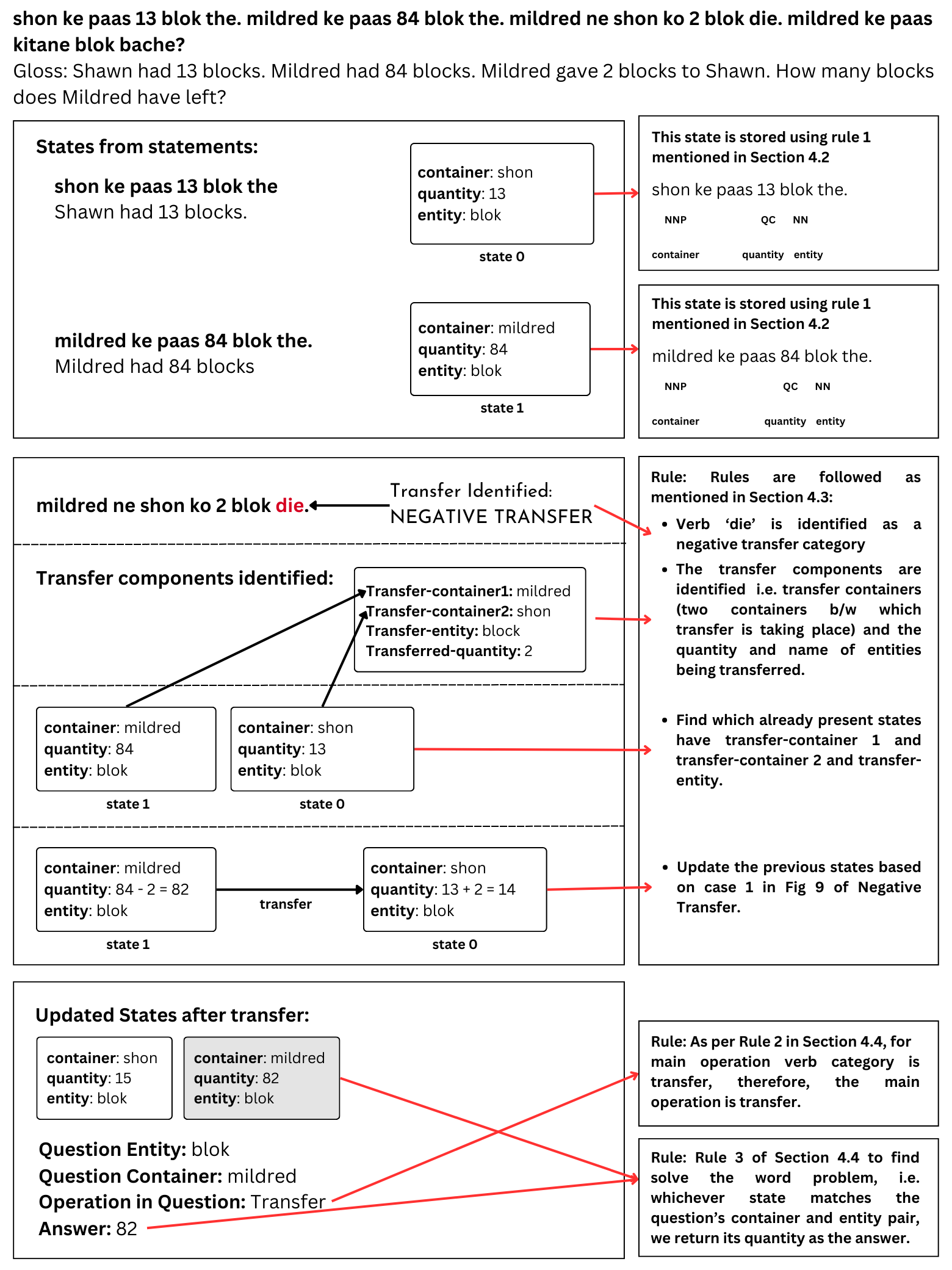}}}
  \caption{Example of solving word problem using the transfer (negative) verb category}
  \label{fig:transfer-verb-eg}
\end{figure*}

\bibliography{anthology,custom}

\onecolumn
\appendix
\section{Appendix} \label{appendix}

\subsection{Examples of rules used in our Hindi Word Problem Solver}

\subsubsection{Example when Main Operation is Transfer}
\begin{figure*}[h]
  \centering
  {%
\setlength{\fboxsep}{0pt}%
\setlength{\fboxrule}{1pt}%
\fbox{\includegraphics[width=1\linewidth]{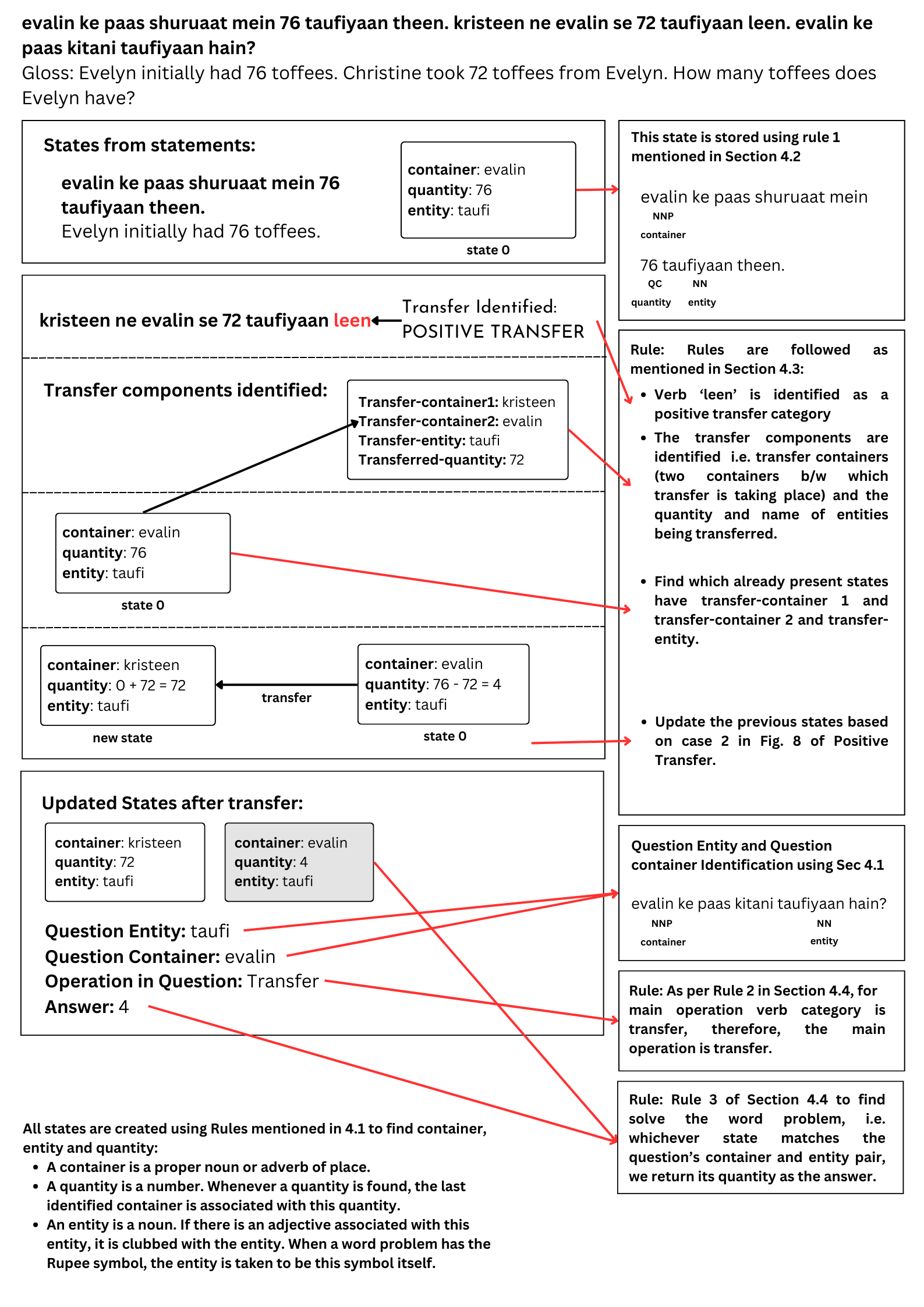}}}
  \caption{Example of solving word problem using negative transfer verb category}
  \label{fig:pos-trans}
\end{figure*}

As mentioned in Section \ref{subsec:answer}, the main operation is `transfer' when a transfer verb category is encountered in the word problem. The transfer verb category may have two types: Positive and negative transfer. An example of Negative Transfer is covered as part of Figure \ref{fig:transfer-verb-eg}. Figure \ref{fig:pos-trans} illustrates an example of how a problem with positive transfer verb category is solved by the solver.

\subsubsection{Examples when the Main operation is Negative}
\begin{figure*}[h]
  \centering
  {%
\setlength{\fboxsep}{0pt}%
\setlength{\fboxrule}{1pt}%
\fbox{\includegraphics[width=1\linewidth]{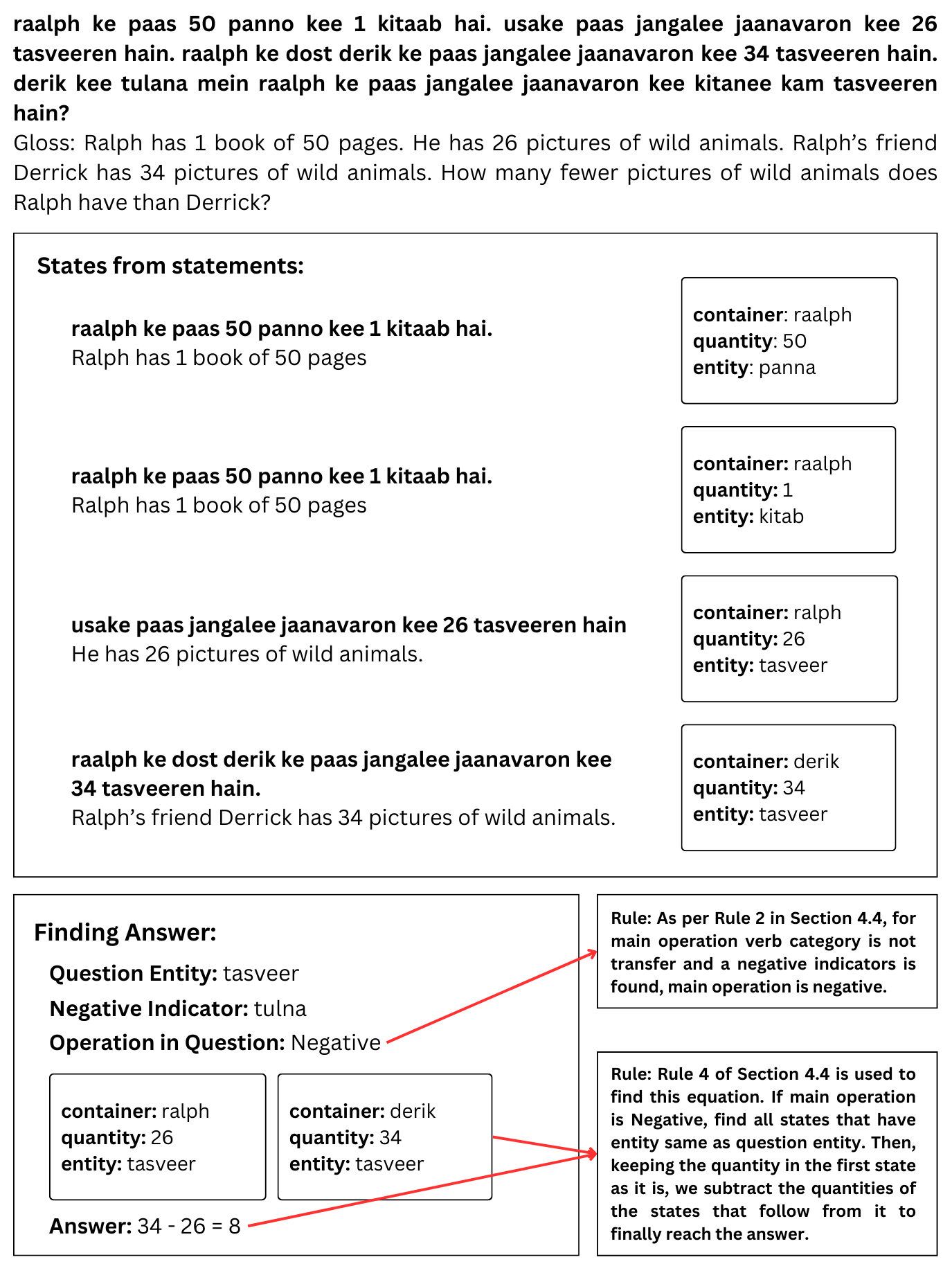}}}
  \caption{Example of solving word problem with negative indicator}
  \label{fig:neg-indi}
\end{figure*}

Section \ref{subsec:answer} states that the main operation is Negative when no transfer verb category is found, and a negative indicator is present in question. Figure \ref{fig:neg-indi} states an example of the same.

\subsubsection{Examples of rules mentioned in Section \ref{subsec:more-rules}}
\begin{itemize}
    \item If the final answer calculated by the solver is negative, we return its absolute value. Example in Figure \ref{fig:abs}.
    \begin{figure*}[h]
  \centering
  {%
\setlength{\fboxsep}{0pt}%
\setlength{\fboxrule}{1pt}%
\fbox{\includegraphics[width=1\linewidth]{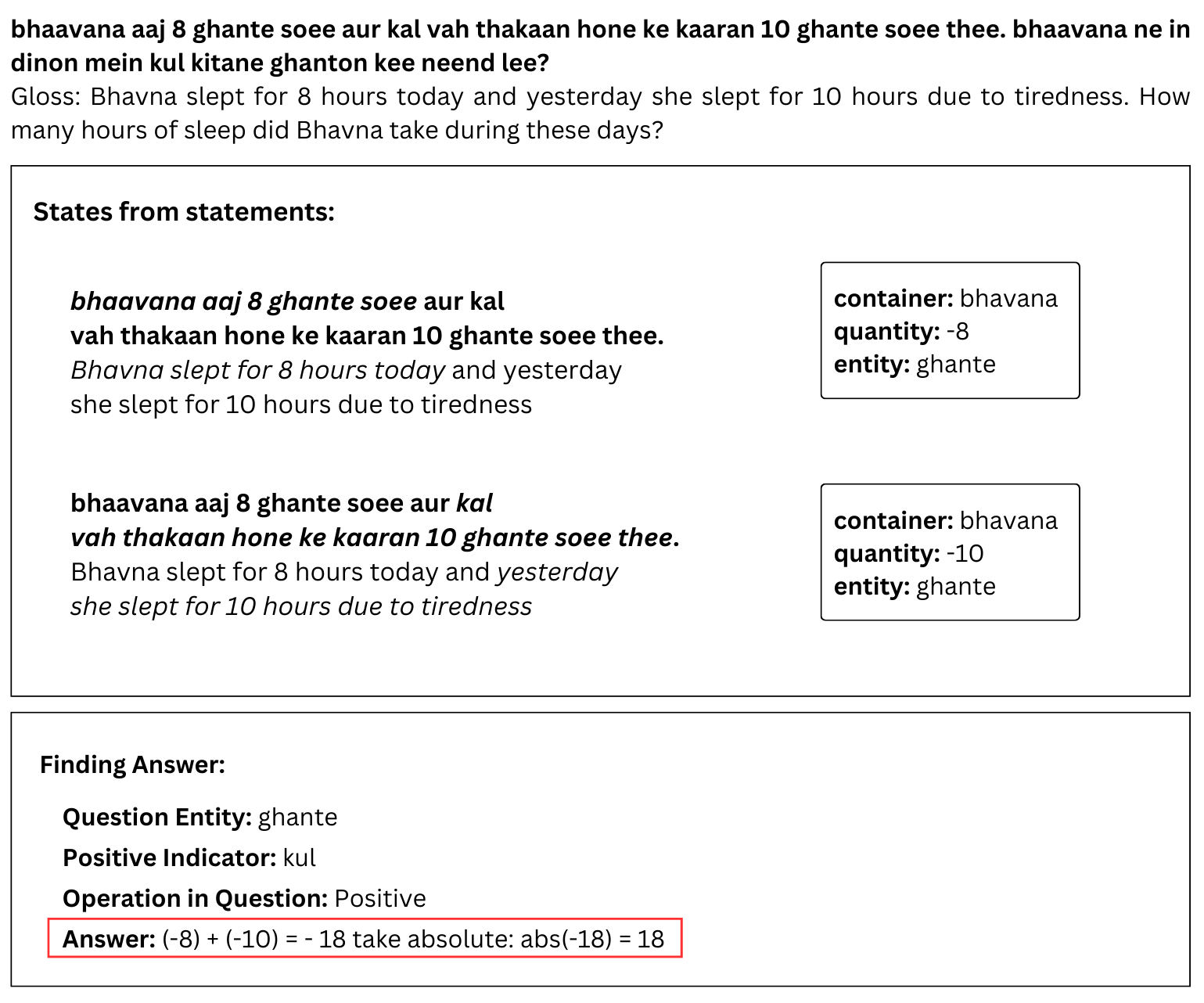}}}
  \caption{Example of solving word problem with negative indicator}
  \label{fig:abs}
\end{figure*}
    \item If the entity in the word problem is found to be one of ‘paisa’, ‘keemat’, ‘laagat’, and ‘rupay’, we change it to the Rupee symbol. Example in Figure \ref{fig:rupee}.
        \begin{figure*}[h]
  \centering
  {%
\setlength{\fboxsep}{0pt}%
\setlength{\fboxrule}{1pt}%
\fbox{\includegraphics[width=1\linewidth]{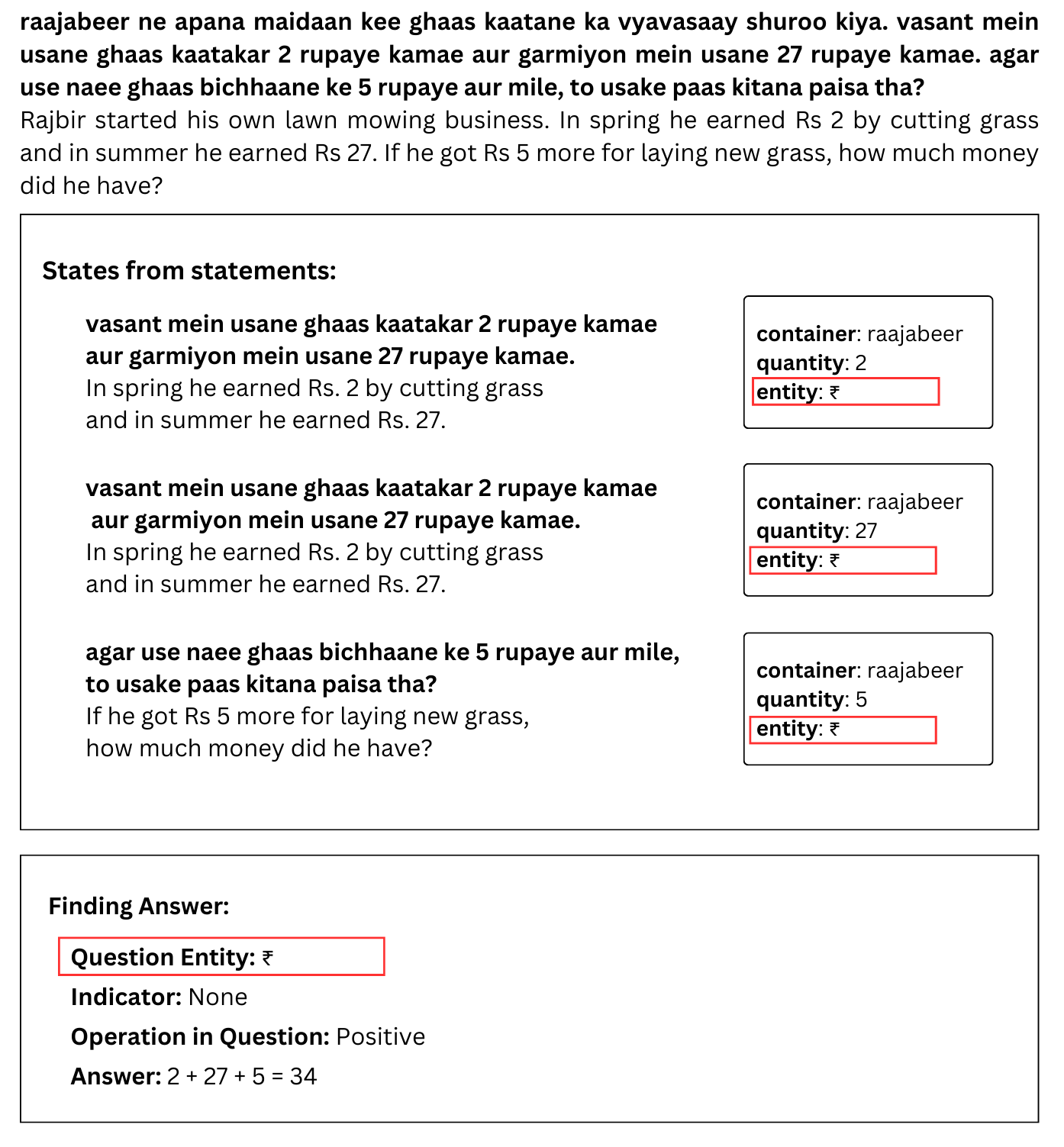}}}
  \caption{Example of solving word problem with negative indicator}
  \label{fig:rupee}
\end{figure*}
    \item If a quantity is found without an entity or container, we retain the same entity and container from the last state and create a new state with the quantity found. Example in Figure \ref{fig:not-found}.
    \begin{figure*}[h]
  \centering
  {%
\setlength{\fboxsep}{0pt}%
\setlength{\fboxrule}{1pt}%
\fbox{\includegraphics[width=1\linewidth]{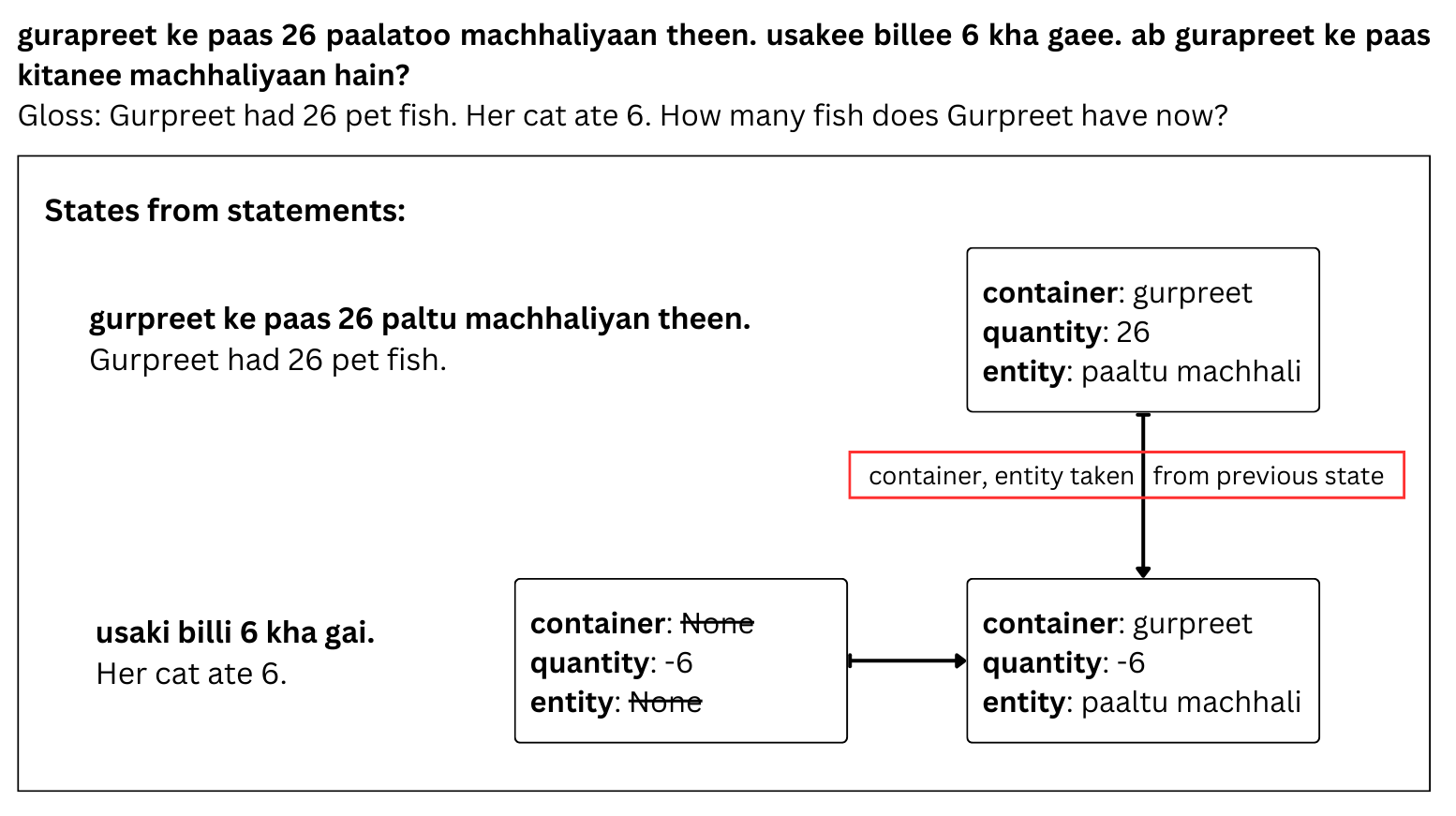}}}
  \caption{Example of solving word problem with negative indicator}
  \label{fig:not-found}
\end{figure*}
    \item If the entity in question is not found in states, we assume the entity of the first state to be the entity in question and perform the steps of finding the answer. Example in Figure \ref{fig:question-entity}.
     \begin{figure*}[h]
  \centering
  {%
\setlength{\fboxsep}{0pt}%
\setlength{\fboxrule}{1pt}%
\fbox{\includegraphics[width=1\linewidth]{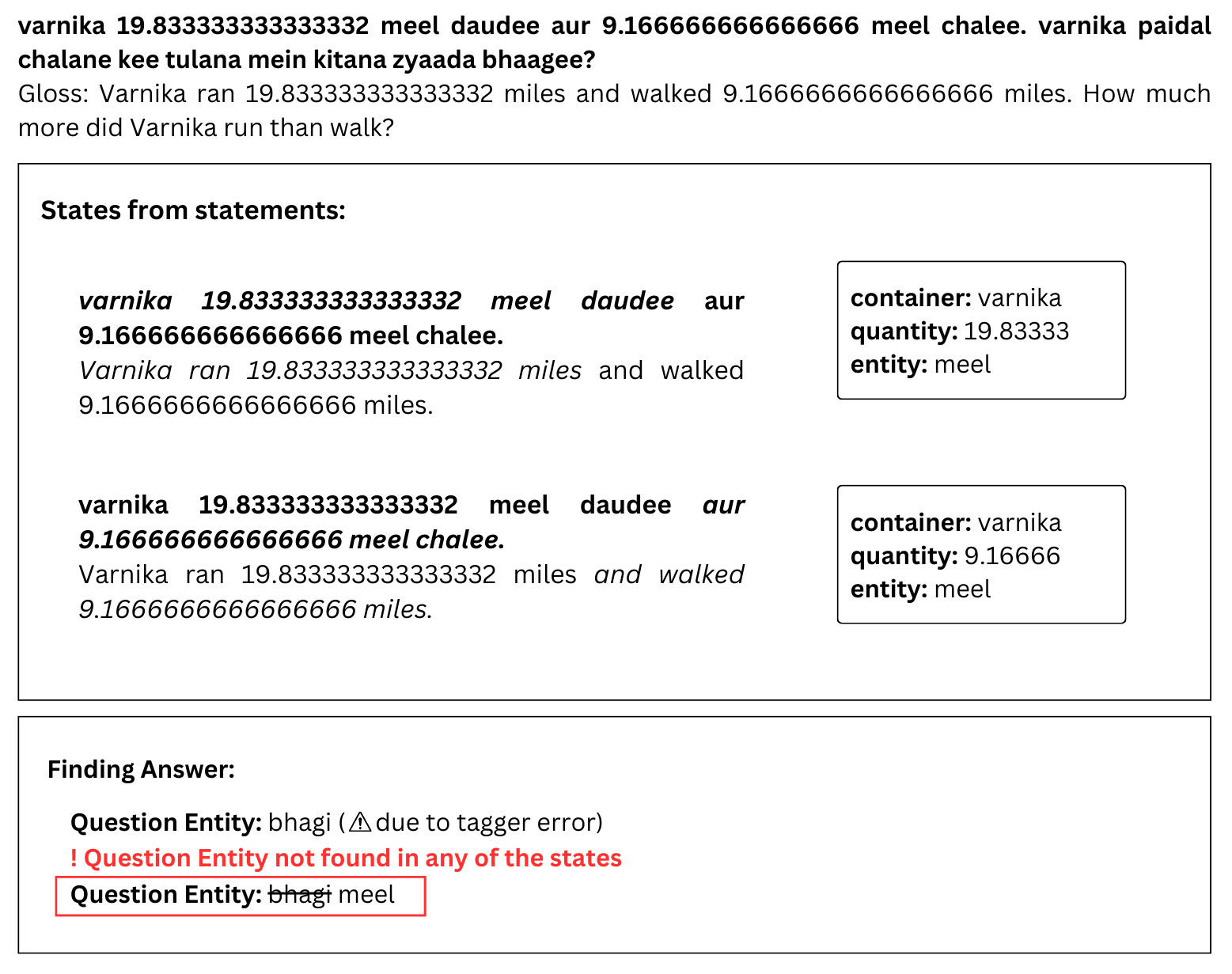}}}
  \caption{Example of solving word problem with negative indicator}
  \label{fig:question-entity}
\end{figure*}
\end{itemize}

\subsection{Examples of Errors made by Solver}
Table \ref{table:solver-errors} gives examples of errors made by the rule-based solver.

\begin{table*}[h]
	\centering
	\begin{tabular}{|p{3cm}|p{12cm}|} 
		\hline
		\textbf{Error Category} & \textbf{Example} \\
		\hline
		Irrelevant Information & raam is maheene 11 kriket ke maich dekhane gaya. vah pichhale maheene 17 maich dekhane gaya tha aur agale maheene 16 maich dekhane jaaega. vah ab tak kul kitane maich dekh chuka hai?
    \\ & Gloss: Ram went to watch 11 cricket matches this month. He went to watch 17 matches last month and next month he will go to watch 16 matches. How many matches has he watched till now?
    \\ & \textit{Error: Solver returns answer as X=11+17+16. 16 matches that Ram will see next month is irrelevant to the question being asked in the word problem.}  \\
   		\hline
		Error in entity/container/action identification & shurooaat mein jen ke paas 87 kele the. 7 1 ghode dvaara khae gae. ant mein jen ke paas kitane kele bache? \\ & Gloss: Initially Jane had 87 bananas. 7 were eaten by 1 horse. How many bananas are left with Jane at the end?
    \\ & \textit{Error: 7 is not mapped to `kele' by the solver and is therefore missed in calculation.}  \\
\hline
Set Completion & 4 bachchon, 2 karmachaariyon aur 3 adhyaapakon ka 1 samooh chidiyaaghar ja raha hai. chidiyaaghar kitane log ja rahe hain?
    \\ & Gloss: 1 group consisting of 4 children, 2 staff and 3 teachers is going to zoo. How many people are going to the zoo?
    \\ & \textit{Error: Here, bacche (children), karmachaari (staff) and adhyaapak (teachers) form a set - log (people), which solver is not capable of identifying.}  \\
		 \hline
   Parsing Errors & evalin ke paas shuruaat mein 76 taaaifiyaan theen. kristeen ne evalin ko 72 taaaifiyaan deen. evalin ke paas kitanee taaaifiyaan hain?
    \\ & Gloss: Evelyn initially had 76 candies. Christine gave 72 candies to Evelyn. How many candies does Evelyn have?
    \\ & \textit{Error: `taaaifiyaan' gets tagged as VM i.e. verb in first statement and is missed from calculation.}  \\
		\hline
	\end{tabular}
	\caption{Examples of erroneous cases}
	\label{table:solver-errors}
\end{table*}


\end{document}